\documentclass[letterpaper, 10 pt, conference]{ieeeconf}  % Comment this line out
\IEEEoverridecommandlockouts                              % This command is only
\usepackage{multirow}
\usepackage{graphicx}
\usepackage{algorithm}
\usepackage{algorithmic}

\title{\LARGE \bf
\textit{dynnode2vec}: Scalable Dynamic Network Embedding
}

\author{Sedigheh Mahdavi$^{1}$, Shima Khoshraftar$^{1}$ and Aijun An$^{1}$% <-this % stops a space
\thanks{$^{1}$Department of Electrical Engineering and Computer Science,
        York University, Toronto, Canada
        {\tt\small \{smahdavi, khoshraf, aan\}@cse.yorku.ca}}%
}

\begin{document}

\maketitle
\thispagestyle{empty}
\pagestyle{empty}

\begin{abstract}
Network representation learning in low dimensional vector space has attracted considerable attention in both academic and industrial domains. Most real-world networks are dynamic with addition/deletion of nodes and edges. The existing graph embedding methods are designed for static networks and they cannot capture evolving  patterns in a large dynamic network. In this paper, we propose a dynamic embedding method, \textit{dynnode2vec}, based on the well-known graph embedding method \textit{node2vec}. \textit{Node2vec} is a random walk based embedding method for static networks.
Applying static network embedding in dynamic settings has two crucial problems: 1) Generating random walks for every time step is time consuming 2) Embedding vector spaces in each timestamp are different. In order to tackle these challenges, \textit{dynnode2vec} uses evolving random walks and initializes the current graph embedding with previous embedding vectors. We demonstrate the advantages of the proposed dynamic network embedding by conducting empirical evaluations on several large dynamic network datasets.
%new embedding vectors in the current 
%as initial weights in 
%Since random-walk based graph embedding are able to handle large static networks, they are potentional  
\end{abstract}
%\begin{IEEEkeywords}
%\textbf{\textit{Keywords}}-Dynamic networks, Graph embedding , Network embedding, Representation learning
%\end{IEEEkeywords}
\section{Introduction}
In the last few decades, graph embedding methods have achieved remarkable success in the analysis of large networks. The basic aim is to represent nodes of a large, high-dimensional graph in low-dimensional vectors that preserve the neighbourhood information of the graph. Several static graph embedding algorithms \cite{cao2015grarep,tang2015line,grover2016node2vec,perozzi2014deepwalk,cao2016deep,chang2015heterogeneous,goyal2018graph,kipf2016variational} have been developed for a variety of machine learning tasks such as visualization \cite{maaten2008visualizing}, node classification \cite{bhagat2011node}, link prediction \cite{liben2007link}, and recommendation \cite{yu2014personalized}. Random walk and edge sampling graph embedding approaches \cite{tang2015line,grover2016node2vec,perozzi2014deepwalk} have a remarkable performance in large networks which contain more than thousand nodes and edges. 

Large real-world networks evolve naturally over time, i.e., nodes and edges appear or disappear, or edges change. For example, in co-authorship networks new edges may emerge (new collaborations may be formed) or new nodes can be added (new authors) and in social networks, users may delete friends (delete edges) or some users may leave the network (delete nodes). The static graph embedding methods are not capable of dealing with the critical challenge involved in dynamic networks. The disadvantage of using static embedding methods at each timestamp independently are as follows. First, embedding vectors for each timestamp are in different spaces. Second, learning embedding vectors separately is a time-consuming process.

In this paper, we propose \textit{dynnode2vec}, a scalable dynamic network embedding for large evolving networks.  In order to handle dynamic networks, \textit{dynnode2vec} modifies the well-known static embedding method, \textit{node2vec} by employing the previous learned embedding vectors as initials weights for the skip-gram model. This is motivated by dynamic language models, especially dynamic Skip-gram models \cite{bamler2017dynamic,kim2014temporal}. In addition, we utilizes evolving random walks for updating the trained skip-gram from previous timestamp. The evloving random walks are only generated for nodes that have changed in consecutive times. As random walk generation is the time consuming part of graph embeddings, we are able to significantly reduce the running time. Our main contributions in this paper are:
\begin{itemize}
    \item We develop a dynamic embedding method \textit{dynnode2vec} that captures evolving patterns in large dynamic networks
    \item \textit{dynnode2vec} is a fast and accurate method for dynamic graph embedding
    \item We evaluate the performance of our method in variety of tasks including link prediction, node classification and anomaly detection on large real-world graphs 
\end{itemize}

\section{\textit{Dynnode2vec}: Scalable dynamic network embedding}
We represent a dynamic network $G$ as a sequence of graphs $G_1,G_2,\dots,G_T$ from timestamps $1$ to $T$. Each graph at time $t$ is represented by $G_t=(V_t,E_t)$ where $V_t$ and $E_t$ are the vertices and edges of the graph respectively. Our goal is to modify the static \textit{node2vec} embedding method for learning representation of a dynamic network. We begin with a brief description of the static \textit{node2vec} embedding method. Then, we explain the steps of \textit{dynnode2vec} which are summarized in Algorithm \ref{alg1}.\\
%and Dynnode2vec extend \textit{node2vec} for feature learning in dynamic networks.
\textit{node2vec} is an extension of \textit{deepwalk} algorithm \cite{perozzi2014deepwalk} and  has two main subcomponents; \textit{node2vec} random walk and Skip-gram model. \textit{node2vec} random walk is a flexible random walk method which samples neighbourhoods of a source node by Breadth-first Sampling (BFS) and Depth-first Sampling (DFS). \textit{node2vec} first generates a corpus by sampling a number of random walks $\gamma$ of length \textit{t} starting at each vertex. Then, the Skip-gram uses these random walks to learn the representation vector for each node. 
%shows the details of Dynnode2vec . 

\subsection{Description of \textit{dynnode2vec} steps}

The main challenge of modification of the static \textit{node2vec} is how to learn embedding at time $t$ by updating embedding vectors in time step $t-1$. For a dynamic network $G={G_1,G_2,\dots,G_T}$, we run the static \textit{node2vec} for the first graph $G_1$ separately, extract the embedding vectors and keep the structure of trained Skip-gram for the next timestamp. For all other subsequent timestamps $2,\dots,T$, the following steps are performed between two consecutive timestamps $t$ and $t+1$. 

\subsubsection{Evolving Walk generation}
%To cope with fast structure evolution in dynamic networks, we update the embedding vectors from timestamp $t-1$ to timestamp $t$ according to the evolving nodes.
In the static \textit{node2vec}, random walks are generated independently for each timestamp for all nodes which is very time-consuming process. In \textit{dynnode2vec}, we generate an effective set of random walks for only evolving nodes instead of generating random walks for all nodes in the current timestamp. Therefore, new random walks from changed regions in the graph $G_t$ can efficiently update the embedding vectors according to temporal evolution of networks over time. 
Assume the structure evolution of the graph $G_t$ from $G_{t-1}$ includes sets of new edges/nodes that are added ($E_{add}/V_{add}$), deleted ($E_{del}/V_{del}$) and their weights have changed ($E_{change}$). The evolving nodes in the timestamp $t$ are defined as follows:
\begin{equation}
\Delta V_t = V_{add}  \cup \{v_i \in V_t | \exists e_i =(v_i,v_j)\in (E_{add} \cup E_{del})\}
\end{equation}
%addition/deletion of nodes/edges from the graph $G_{t-1}=(V_{t-1},E_{t-1})$. The evolving nodes is denoted by $\Delta V_t$  in the timestamp $t$ which can achieved
%
Generating evolving random walks is fast and time-efficient since dynamic networks are evolved gradually and neighbourhoods of most nodes are kept unchanged. However, in the worst case, evolving nodes include all nodes as all nodes have changed in the timestamp $t$.

\subsubsection{Dynamic Skip-gram model} 
In natural language processing domain, several dynamic word embedding \cite{kaji2017incremental,may2017streaming,bamler2017dynamic,kim2014temporal} have been proposed to track word evolution such as new words are created (internet), and some words ‘die out’ throughout time. In \cite{kim2014temporal}, for learning the representation vectors in the timestamp $t$ they train the Skip-gram by initializing word vectors obtained from previous timestamp $t-1$. In this dynamic Skip-gram model, the vocabulary set is updated and the Skip-gram is retrained by new documents in timestamp $t$. In network embedding domains, DynGEM \cite{goyal2018dyngem} incrementally learns embedding vectors for the timestamp $t$ by using embedding vectors from the timestamp $t-1$ as initial weights for SDNE atuoencoder in a dynamic network.

\textit{dynnode2vec} also takes advantage of dynamic Skip-gram model for obtaining embedding vectors at time $t$ and uses the pre-trained Skip-gram model (Skip-gram$_{t-1}$) as initial weight for (Skip-gram$_{t}$). In order to do that, first the vocabulary set of Skip-gram$_{t}$ is updated according to new evolving walks. Then, Skip-gram$_{t}$ is trained by new evolving walks generated on the evolving nodes.

\begin{algorithm}
%\floatpagestyle{empty}
\caption{:\textbf{Algorithm: Dynnode2vec}}
\label{alg1}
\begin{algorithmic} [1]
\STATE \textbf{Input}: Graphs $G={G_1,G_2,\dots,G_T}$
\STATE \textbf{Output}: Embedding vectors $Z_1,Z_2,\dots,Z_T$
%\STATE
\STATE Run static \textit{node2vec} for the Graph $G_1$
\FOR {$t =2$ to $N$}
\STATE Find a set of evolving nodes,$\Delta V_t$,
\STATE Sample new random walks ($Walk_n$) for $\Delta V_t$
%\STATE Update vocabulary in $Skip_{t-1}$ according new random walks
\STATE Train Skip-Gram $Skip_{t}$ with $Walk_n$ and obtain $Z_t$
%\STATE 
\ENDFOR
%\STATE $NFC=NFC+NP$;
%\ENDWHILE
\end{algorithmic}
\end{algorithm}
%controlled by two parameters \textit{p} and \textit{q}.
%Skip-gram architectures 

\section{EXPERIMENTS}

\subsection{Datasets}
The performance of \textit{dynnode2vec} is evaluated on following datasets. 
\begin{itemize}
    \item Hep-th \cite{hepth}: This dataset is the coauthorship network of researchers in High energy physics theory conference. Hep-th has $34k$ nodes, $421k$ edges divided into 60 graphs.
    \item Autonomous Systems (AS) \cite{as}: AS is built from logs of the BGP (Border Gateway Protocol) which shows the communication between users. Number of nodes, edges and time steps are $6k$, $13k$ and 100 respectively.
    \item Enron \cite{enron}: Enron is the email communication network between Enron company employees. It has $87k$ nodes and $1.1M$ edges over $175$ months.
    \item StackOverflow (St-Ov) \cite{stov}: This dataset is derived from question and answers in Math Overflow website. Each edge shows users interactions. This dataset contains $14k$ nodes and $195k$ edges over 2350 days. We divided the datastream into 58 graphs.
    \item dblp \cite{dblp}: This is the coauthorship network dataset among researchers of different fields. It consists of $90k$ nodes and $749k$ edges over 18 years. Each node in dblp has one of the two class labels, database and data mining (VLDB, SIGMOD, PODS, ICDE, EDBT, SIGKDD, ICDM, DASFAA, SSDBM, CIKM, PAKDD, PKDD, SDM and DEXA) and computer vision and pattern recognition (CVPR, ICCV, ICIP, ICPR, ECCV, ICME and ACM-MM).
    
    \item Darpa \cite{darpa}: Darpa is a dataset consisting of communications between source IPs and destination IPs. This dataset contains different attacks between IPs. We used a subset of darpa consisting of $12k$ nodes and $22k$ edges over $100$ hours. 
\end{itemize}

\subsection{Baselines}
We compared the performance of our method against the following existing methods:
\begin{itemize}
    \item DeepWalk\cite{perozzi2014deepwalk}: This is the first node embedding method based on random walks. DeepWalk uses Skip-gram model and uniform random walks to learn the neighborhood structure of the graph. 
    \item \textit{node2vec}\cite{grover2016node2vec}:  This method learns node representation in networks by preserving network neighborhood of nodes. It explores the neighborhood of a node by generating DFS and BFS random walks for that node.
    \item DynGEM \cite{goyal2018dyngem}: This approach is a stable dynamic node embedding method that works for growing dynamic graphs. It incrementally builds the embedding vectors at each time using the embedding vectors from previous time.  In dynamic networks with a large number of nodes like Enron and DBLP, the memory requirement of Dyngem signicantly increases. Therefore, we are only able to run it on three datasets; AS, Hep-th, and St-Ov.

\end{itemize}

\subsection{Link Prediction}
Link prediction is an important application of graph embeddings. In this task, future edges are predicted given the previous edges.  We consider the link prediction as a classification task similar to \cite{grover2016node2vec}. For example, if we have a sequence of graphs $G_0, G_1, ...,G_t$, we predict edges at time t using edges from time $0$ to ${t-1}$. For instance,
edges of $G_1$ are predicted using the positive and negative edges of $G_0$. For $G_2$, we use edges from $G_0$ and $G_1$. We do the same for all the graphs at future time points. \\
\textit{Edge embedding}. The embedding of an edge $(u,v)$ can be computed using embedding vectors of nodes $u$ and $v$. In the literature, different operators are applied on node embedding vectors to compute edge embeddings including Weighted-L1, Weighted-L2, Hadamard and average as defined in \cite{grover2016node2vec}.\\
We report the link prediction results for all the four aforementioned operators in Table \ref{tab1}. The results are the average AUC (Area Under Curve) with a grid search over $p,q \in \{0.5,1,2,4\}$. We evaluated all methods on four datasets: AS, Hep-th, Enron and St-Ov. The results show that \textit{dynnode2vec} outperforms baselines in almost all datasets. Among different operators, \textit{dynnode2vec} has the best performance with Hadamard operator and the worst performance with average operator.

\begin{table}[]
\begin{center}
\caption{Link prediction results using four operators a) Weighted-L1 b) Weighted-L2 c) Hadamard d) Average}
\label{tab1}
\begin{tabular}{clllll}
\multicolumn{1}{l}{\textbf{Op}} & \multicolumn{1}{c}{\textbf{Algorithm}} & \multicolumn{4}{c}{\textbf{Dataset}}                                                                                                 \\
\multicolumn{1}{l|}{}           & \multicolumn{1}{l|}{}                  & \multicolumn{1}{l|}{AS}              & \multicolumn{1}{l|}{Hep-th}          & \multicolumn{1}{l|}{Enron}           & St-Ov           \\ \hline
\multicolumn{1}{c|}{A}          & \multicolumn{1}{l|}{DeepWalk}          & \multicolumn{1}{l|}{0.957}           & \multicolumn{1}{l|}{0.9887}          & \multicolumn{1}{l|}{0.8489}          & 0.5595          \\
\multicolumn{1}{c|}{}           & \multicolumn{1}{l|}{node2vec}          & \multicolumn{1}{l|}{0.9554}          & \multicolumn{1}{l|}{0.9893}          & \multicolumn{1}{l|}{0.8533}          & 0.5617          \\
\multicolumn{1}{c|}{}           & \multicolumn{1}{l|}{DynGEM}            & \multicolumn{1}{l|}{0.7855}          & \multicolumn{1}{l|}{0.6246}          & \multicolumn{1}{l|}{-}               & 0.6242          \\
\multicolumn{1}{c|}{}           & \multicolumn{1}{l|}{dynnode2vec}       & \multicolumn{1}{l|}{\textbf{0.9625}} & \multicolumn{1}{l|}{\textbf{0.996}}  & \multicolumn{1}{l|}{\textbf{0.8922}} & \textbf{0.66}   \\ \hline
\multicolumn{1}{c|}{B}          & \multicolumn{1}{l|}{DeepWalk}          & \multicolumn{1}{l|}{0.9577}          & \multicolumn{1}{l|}{0.9882}          & \multicolumn{1}{l|}{0.805}           & 0.5561          \\
\multicolumn{1}{c|}{}           & \multicolumn{1}{l|}{node2vec}          & \multicolumn{1}{l|}{0.9561}          & \multicolumn{1}{l|}{0.9894}          & \multicolumn{1}{l|}{0.8649}          & 0.5608          \\
\multicolumn{1}{c|}{}           & \multicolumn{1}{l|}{DynGEM}            & \multicolumn{1}{l|}{0.7701}          & \multicolumn{1}{l|}{0.6161}          & \multicolumn{1}{l|}{-}               & 0.6246          \\
\multicolumn{1}{c|}{}           & \multicolumn{1}{l|}{dynnode2vec}       & \multicolumn{1}{l|}{\textbf{0.9635}} & \multicolumn{1}{l|}{\textbf{0.9977}} & \multicolumn{1}{l|}{\textbf{0.8981}} & \textbf{0.6698} \\ \hline
\multicolumn{1}{c|}{C}          & \multicolumn{1}{l|}{DeepWalk}          & \multicolumn{1}{l|}{0.888}           & \multicolumn{1}{l|}{0.9749}          & \multicolumn{1}{l|}{0.8167}          & 0.7456          \\
\multicolumn{1}{c|}{}           & \multicolumn{1}{l|}{node2vec}          & \multicolumn{1}{l|}{0.8935}          & \multicolumn{1}{l|}{0.9762}          & \multicolumn{1}{l|}{0.8236}          & 0.7478          \\
\multicolumn{1}{c|}{}           & \multicolumn{1}{l|}{DynGEM}            & \multicolumn{1}{l|}{0.8005}          & \multicolumn{1}{l|}{0.5882}          & \multicolumn{1}{l|}{-}               & 0.6387          \\
\multicolumn{1}{c|}{}           & \multicolumn{1}{l|}{dynnode2vec}       & \multicolumn{1}{l|}{\textbf{0.9512}} & \multicolumn{1}{l|}{\textbf{0.9975}} & \multicolumn{1}{l|}{\textbf{0.9012}} & \textbf{0.8886} \\ \hline
\multicolumn{1}{c|}{D}          & \multicolumn{1}{l|}{DeepWalk}          & \multicolumn{1}{l|}{0.6197}          & \multicolumn{1}{l|}{0.5571}          & \multicolumn{1}{l|}{0.5271}          & 0.5182          \\
\multicolumn{1}{c|}{}           & \multicolumn{1}{l|}{node2vec}          & \multicolumn{1}{l|}{0.6258}          & \multicolumn{1}{l|}{0.5577}          & \multicolumn{1}{l|}{0.5083}          & 0.5149          \\
\multicolumn{1}{c|}{}           & \multicolumn{1}{l|}{DynGEM}            & \multicolumn{1}{l|}{\textbf{0.7949}} & \multicolumn{1}{l|}{0.555}           & \multicolumn{1}{l|}{-}               & 0.6175          \\
\multicolumn{1}{c|}{}           & \multicolumn{1}{l|}{dynnode2vec}       & \multicolumn{1}{l|}{0.7718}          & \multicolumn{1}{l|}{\textbf{0.6269}} & \multicolumn{1}{l|}{\textbf{0.6147}} & \textbf{0.6824} \\ \hline
\multicolumn{2}{l|}{node2vec settings (p,q)}                             & \multicolumn{1}{l|}{(0.5,1)}         & \multicolumn{1}{l|}{(0.5,1)}         & \multicolumn{1}{l|}{(0.5,1)}         & (1,2)          
\end{tabular}

\end{center}
\end{table}

\subsection{Node Classification}
Another application of node embeddings is in node classification. In supervised classification, nodes have class labels in the dataset. We used logistic regression as the classification method. Similar to link prediction, node embeddings of previous time points are used in predicting class labels of future time points. We compared the results of our method with baselines on dblp dataset. The Micro-$F_1$ and Macro-$F_1$ results are reported in Table \ref{tab2}. Our result are better than other baselines in terms of Micro-$F_1$ scores.

\begin{table}[]
\begin{center}
\caption{Node classification results}
\label{tab2}
\begin{tabular}{lll}
\textbf{Metric}                                & \multicolumn{1}{c}{\textbf{Algorithm}} & \multicolumn{1}{c}{\textbf{Dataset}} \\
\multicolumn{1}{l|}{}                          & \multicolumn{1}{l|}{}                  & dblp                                 \\ \hline
\multicolumn{1}{l|}{\multirow{3}{*}{Micro-F1}} & \multicolumn{1}{l|}{DeepWalk}          & 0.5337                               \\
\multicolumn{1}{l|}{}                          & \multicolumn{1}{l|}{node2vec}          & 0.5272                               \\
\multicolumn{1}{l|}{}                          & \multicolumn{1}{l|}{dynnode2vec}       & \textbf{0.5415}                      \\ \hline
\multicolumn{1}{l|}{\multirow{3}{*}{Macro-F1}} & \multicolumn{1}{l|}{DeepWalk}          & \textbf{0.4141}                      \\
\multicolumn{1}{l|}{}                          & \multicolumn{1}{l|}{node2vec}          & 0.3847                               \\
\multicolumn{1}{l|}{}                          & \multicolumn{1}{l|}{dynnode2vec}       & 0.4056                               \\ \hline
\multicolumn{2}{l|}{node2vec settings (p,q)}                                            & (1,2)                               
\end{tabular}

\end{center}
\end{table}

\subsection{Anomaly Detection}
\begin{figure}
    \centering
    \includegraphics[scale=0.8]{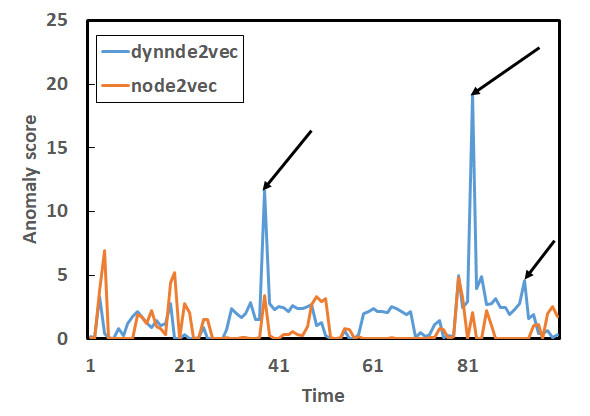}
    \caption{Anomaly detection results}
    \label{fig1}
\end{figure}

Anomalies are any deviations from normal behavior. There are different categories of anomaly detection methods in dynamic settings including detecting anomalous nodes, edges and change detection. Inspired by DynGEM, we used \textit{dynnode2vec} to detect changes in the dynamic network. We computed the norm of the differences between embedding vectors of common nodes at consecutive time points. High values for the norm signals a significant change in the structure of the graph. We applied \textit{dynnode2vec} and node2vec on a subset of Darpa dataset\cite{darpa}. Figure \ref{fig1} shows that there are three major peaks in the dynamic node2vec curve. These spikes correspond to time points that three important attacks occurred in the dataset. 

\subsection{Effects of evolving walk generation}
In this work, we only generate walks for the nodes in the graph that have changed compared to previous time point. This leads to a significant speedup in \textit{dynnod2vec} running time. In order to show the time efficiency of \textit{dynnod2vec}, we compared the running time of \textit{dynnod2vec} with two methods: node2vec and \textit{dynnod2vec} version denoted by \textit{dynnod2vec-all} that samples walks for all the nodes. The comparison results on three datasets: AS (the first fifty time steps), Hep-th and Enron (the first forty time steps) are shown in Table \ref{tab3}. \textit{dynnod2vec} was faster than other methods. All experiments are performed on a windows X-64 with 7 cores, 64 GB RAM and a clock speed of 3.6 GHz.\\
Additionally in terms of accuracy, we compared the AUC scores of \textit{dynnod2vec} with \textit{dynnod2vec-all}. Table \ref{tab4} indicates the AUC scores in link prediction task. In most cases the AUC for these two methods are not significantly different. 

\begin{table}[]
\begin{center}
\caption{Running time comparison}
\label{tab3}
\begin{tabular}{l|l|l}
                & AS        & Hep-th    \\ \hline
node2vec        & 23.22 min & 14.42 min \\
dynnode2vec-all & 23.10 min & 13.52 min \\
dynnode2vec     & 4.49 min  & 1.25 min 
\end{tabular}
\end{center}
\end{table}

\begin{table}[]
\begin{center}
\caption{Comparison of dynnode2vec vs dynnode2vec-all}
\label{tab4}
\begin{tabular}{clllll}
\textbf{OP}            & \multicolumn{1}{c}{\textbf{Algorithm}} & \multicolumn{4}{c}{\textbf{Dataset}}                                                                \\
\multicolumn{1}{l|}{}  & \multicolumn{1}{l|}{}                  & \multicolumn{1}{l|}{AS}      & \multicolumn{1}{l|}{Hep-th}  & \multicolumn{1}{l|}{Enron}   & St-Ov  \\ \hline
\multicolumn{1}{c|}{a} & \multicolumn{1}{l|}{dynnode2vec-all}   & \multicolumn{1}{l|}{0.9799}  & \multicolumn{1}{l|}{0.9973}  & \multicolumn{1}{l|}{0.9114}  & 0.6674 \\
\multicolumn{1}{c|}{}  & \multicolumn{1}{l|}{dynnode2vec}       & \multicolumn{1}{l|}{0.9625}  & \multicolumn{1}{l|}{0.996}   & \multicolumn{1}{l|}{0.8922}  & 0.66   \\ \hline
\multicolumn{1}{c|}{b} & \multicolumn{1}{l|}{dynnode2vec-all}   & \multicolumn{1}{l|}{0.9803}  & \multicolumn{1}{l|}{0.9976}  & \multicolumn{1}{l|}{0.9042}  & 0.6734 \\
\multicolumn{1}{c|}{}  & \multicolumn{1}{l|}{dynnode2vec}       & \multicolumn{1}{l|}{0.9635}  & \multicolumn{1}{l|}{0.9977}  & \multicolumn{1}{l|}{0.8981}  & 0.6698 \\ \hline
\multicolumn{1}{c|}{c} & \multicolumn{1}{l|}{dynnode2vec-all}   & \multicolumn{1}{l|}{0.9099}  & \multicolumn{1}{l|}{0.9966}  & \multicolumn{1}{l|}{0.9022}  & 0.8825 \\
\multicolumn{1}{c|}{}  & \multicolumn{1}{l|}{dynnode2vec}       & \multicolumn{1}{l|}{0.9512}  & \multicolumn{1}{l|}{0.9975}  & \multicolumn{1}{l|}{0.9012}  & 0.8886 \\ \hline
\multicolumn{1}{c|}{d} & \multicolumn{1}{l|}{dynnode2vec-all}   & \multicolumn{1}{l|}{0.6488}  & \multicolumn{1}{l|}{0.5982}  & \multicolumn{1}{l|}{0.6364}  & 0.6798 \\
\multicolumn{1}{c|}{}  & \multicolumn{1}{l|}{dynnode2vec}       & \multicolumn{1}{l|}{0.7718}  & \multicolumn{1}{l|}{0.6269}  & \multicolumn{1}{l|}{0.6147}  & 0.6824 \\ \hline
\multicolumn{2}{l|}{node2vec setting (p,q)}                     & \multicolumn{1}{l|}{(0.5,1)} & \multicolumn{1}{l|}{(0.5,1)} & \multicolumn{1}{l|}{(0.5,1)} & (1,2) 
\end{tabular}

\end{center}
\end{table}

\section{Conclusion}
In this paper, we propose \textit{dynnode2vec}, a scalable dynamic network embedding that learns representation vectors for dynamic networks. \textit{dynnode2vec} employs the dynamic Skip-gram model and evolving random walks to discover information changes in temporal networks. In the dynamic Skip-gram model, the previous learned embedding vectors are transferred to the next timestamp as initial weights. This results in smooth embedding vectors for graphs over times.
Furthermore, the evolving random walks are generated to efficiently reflect the changes in dynamic graph structure. By only considering subset of random walks, \textit{dynnode2vec} can obtain embedding vectors in notably less time without sacrificing accuracy. Our experiments demonstrate the superiority of \textit{dynnode2vec} as compared with the state-of-the-art embedding methods in various tasks. Future work will investigate using other dynamic Skip-gram models \cite{kaji2017incremental,bamler2017dynamic} for dynamic graph embedding. 

\section{Acknowledgements}
We would like to thank Palash Goyal for kindly sharing the source code for the DynGEM method.

\addtolength{\textheight}{-12cm}   % This command serves to balance the column lengths

%%%%%%%%%%%%%%%%%%%%%%%%%%%%%%%%%%%%%%%%%%%%%%%%%%%%%%%%%%%%%%%%%%%%%%%%%%%%%%%%

%%%%%%%%%%%%%%%%%%%%%%%%%%%%%%%%%%%%%%%%%%%%%%%%%%%%%%%%%%%%%%%%%%%%%%%%%%%%%%%%

\end{document}